\documentclass[letterpaper]{article}
\usepackage{aaai24}
\usepackage{times}
\usepackage{helvet}
\usepackage{courier}
\usepackage[hyphens]{url}
\usepackage{graphicx}
\usepackage{natbib}
\usepackage{caption}
\usepackage{algorithm}
\usepackage{algorithmic}
\usepackage{bm}
\usepackage{multirow}
\usepackage{booktabs}
\usepackage{amsmath}
\usepackage{amssymb}
\usepackage{amsfonts}
\usepackage{xspace}
\usepackage{placeins}
\usepackage{float}
\usepackage{dblfloatfix}

\nocopyright
\title{SPRKD: Effective Knowledge Distillation for Deep Neural Networks via Saddle Region Approximation}

\author{
    Aditya Dewan\textsuperscript{\rm 1}\thanks{Correspondence to: \texttt{adewan2@andrew.cmu.edu}, \texttt{aditya.dewan124@gmail.com}.}\thanks{This work was awarded 2nd Place in Mathematical and Cybersecurity Research from the U.S.~National Security Agency at the 2023 Regeneron International Science and Engineering Fair (ISEF), 4th Place in Robotics and Intelligent Machines at Regeneron ISEF 2023, and one of 8 projects selected by Youth Science Canada to represent Canada at the 2023 Regeneron Science and Engineering Fair in Dallas, Texas.},
    Arjun Yogeswaran\thanks{Equal contribution as mentors. No affiliation.},
    Benjamin Fedoruk\textsuperscript{\rm 2}\footnotemark[3]
}
\affiliations{
    \textsuperscript{\rm 1}Carnegie Mellon University \quad
    \textsuperscript{\rm 2}Ontario Tech University \\
    \texttt{adewan2@andrew.cmu.edu}, \texttt{arjyoges@gmail.com}, \texttt{benjamin.fedoruk@ontariotechu.ca}
}

\begin{document}

\maketitle
\pagestyle{empty}

\begin{abstract}
Modern deep neural networks (DNNs) are potent catalysts for scientific and industrial impact, yet excessive parameter counts impede deployment in low-compute settings such as hospital equipment and energy infrastructure, elevating required resources and hindering impact \citep{alkhulaifi2021kd}. Predominant knowledge distillation (KD) methods favor replication: smaller students typically mimic teacher output logits, yet empirically yield low basic-task performance, hamper end convergence, inhibit generalization, and act merely as regularization rather than substantive knowledge transfer \citep{cho2019efficacy, stanton2021does, yuan2020revisiting}. We propose Saddle Point Recruitment for Knowledge Distillation (SPRKD), reframing distillation from replication to employing teachers as optimization-curvature and domain proxies and characterizing saddle points as regions of strong further-descent potential through embedding and basin-fractal properties \citep{zhang2021embedding, liao2017theory}. Via Hessian eigenvalue spectral density (ESD), computed with \texttt{PyHessian} \citep{yao2020pyhessian} and \texttt{hessian-eigenthings} \citep{golmant2018hessianeigenthings}, SPRKD identifies low-loss saddle regions for student re-exploration; weak-teacher ensembles are aggregated into an Approximated Saddle Region (ASR), re-parameterized into the student via Transfer Learning by Injection, and approached with exponentially decaying Euclidean transformations, Negative Hessian Eigensteps, and Gaussian perturbations for teacher-boosted descent \citep{jin2017escape, alain2019negative}. On malaria blood smear classification with a 6{,}430-parameter CNN distilled from a weak 25{,}546-parameter teacher, SPRKD reaches 94.8\% validation accuracy, outperforming replication-based Response KD by 24.70 percentage points (McNemar $p = 6.3 \times 10^{-87}$) and matching scratch-trained SGD baselines of the same architecture to statistical equivalence ($p = 1.0$ vs.~Control-S). Across malaria, MNIST, CIFAR-100, and TinyImageNet, SPRKD exceeds scratch-trained baselines by up to 8 percentage points on supplementary benchmarks and Response KD by 24.70 percentage points on malaria, with faster, more complete, and more stable convergence. Multi-teacher weak-ensemble aggregation consistently outperformed single-teacher distillation \citep{asif2020ensemble}. Hessian ESD analysis and 2-D loss landscape visualization across benchmark datasets show wider minima convergence and substantially smaller Hessian trace and spectral radius than Response KD and control students \citep{ghorbani2019hessian, keskar2017large, li2018visualizing}, denoting smoother descent, enhanced noise robustness, and stronger real-world performance. Code, notebooks, and reproduction scripts are available at \url{https://github.com/thetechdude124/SADDLE-POINT-RECRUITMENT-FOR-KNOWLEDGE-DISTILLATION}. These results highlight strong SPRKD impact potential, enabling high-performance model deployment across widespread low-latency and edge pursuits without requiring strong teachers.
\end{abstract}

\section{Introduction}

Modern Deep Neural Networks (DNNs) learn through stochastic optimization of a task-specific objective function (typically cross-entropy between softmax-activated predictions and true labels) via stochastic gradient descent: $\theta_{t+1} = \theta_t - \eta \cdot \nabla \mathcal{L}(\theta_t)$ \citep{domingos2012useful}. For high-performance learning to take place, parameter space $\Theta$ must be sufficiently large to include desired classifiers \citep{domingos2012useful}. As a consequence, modern DNNs are architected with inflated parameter counts and excessive depth, catalyzing high accuracy at the cost of large model size and substantial inference latency \citep{alkhulaifi2021kd}. Large-scale DNN training can require days of compute across many accelerators and incur substantial energy and carbon costs \citep{strubell2019energy}, yet cannot be deployed on many of the problems they were built to solve due to sheer size and latency requirements. Consider intensive care units, where even sub-second inference delays can materially affect acute-care decision making \citep{alkhulaifi2021kd}: deployment must be low latency, high performance, and local to preserve patient privacy. Today, model size forces inference onto the cloud rather than onto edge devices closest to these problems, requiring gigabytes of private user data to be transferred to distant servers before insights can be acted upon. This size and latency overhead inhibits deployment in low-compute, remote, real-time, and mission-critical environments, including hospital ECG equipment, robotics platforms, energy grid monitoring, traffic infrastructure, and the billions of connected edge and IoT devices projected worldwide \citep{ericsson2022iot, alkhulaifi2021kd}.

\paragraph{The Current Solution: Knowledge Distillation.}
Knowledge Distillation (KD) is the presently dominant method of compressing large models into lightweight form factors \citep{hinton2015distilling, gou2021kd, alkhulaifi2021kd}. Through (typically) replicating teacher model output logits, a student network aims to parameterize the same function as the teacher in a compressed architecture by minimizing $\mathcal{L}_{KD} = \sum_{x \in \mathbf{x}} \mathcal{L}_D(s(x), t(x))$, generating functionally equivalent models \citep{hinton2015distilling}. Here $\mathcal{L}_D$ is typically the Kullback-Leibler divergence between softmax-activated student and teacher logits \citep{hinton2015distilling, gou2021kd}.

\paragraph{Where Replication-Based KD Falls Short.}
The reliance of KD approaches on replication empirically enforces an artificial ceiling on student learning capacity, generalization, and performance, effectively inhibiting student models from exceeding teacher performance and establishing teacher strength as the primary factor behind distillation effectiveness (an exception is self-distillation, where both student and teacher are identical) \citep{cho2019efficacy, stanton2021does, furlanello2018born, allenzhu2020ensemble}. KD empirically (i) exhibits low performance on moderately complex tasks such as ImageNet \citep{cho2019efficacy}, (ii) hampers long-term student convergence \citep{stanton2021does}, (iii) requires simultaneous teacher and student inference, doubling training time and resources \citep{stanton2021does, gou2021kd}, and (iv) has been shown to be identical in effect to label-smoothing regularization \citep{yuan2020revisiting}. While valuable for multi-task and regularization benefits, the core compression objective (incorporating teacher knowledge to boost student performance) is left largely unaddressed. Hospital data is not centralized: it is split across dozens of locations and requires expert annotation and regulatory approval to consolidate. Even when sufficient data exist, replication-based KD still requires running both teacher and student simultaneously. This is particularly detrimental in critical settings where strong teacher training is itself infeasible (for example, healthcare DNNs that require extensive expert data annotation under strict regulatory constraints).

\paragraph{Our Approach: Re-Characterizing Distillation.}
It is proposed herein that the solution to current KD limitations involves re-characterizing the fundamental objective of distillation from teacher replication to leveraging teachers as direct proxies for loss landscape curvature information. The goal is not merely a smaller model, but a smaller model that exhibits similar generalization properties to a teacher while \emph{improving} task accuracy beyond what replication allows. More specifically, we employ the intrinsic knowledge and domain information embedded within the loss landscape and optimization process of teacher models to boost student performance and generalization on the given task, rather than solely attempting compression by replication. Teachers provide vetted regions of the loss landscape; students build upon this second-order information while training on the actual task labels. In particular, we use teacher ensembles to approximate regions of the loss landscape with strong potential for further descent upon student re-exploration, concretely accelerating the optimization process beyond student inductive biases.

We achieve this through \emph{saddle points} (SPs): points where $\nabla f(\theta_t) = 0$ and the Hessian $H_f(\theta_t)$ has at least one positive and one negative eigenvalue \citep{dauphin2014identifying}. In Section~\ref{sec:five-tenets} we develop a set of five complementary theoretical and empirical properties of saddle points that make them ideal carriers of teacher knowledge for student models. In Section~\ref{sec:method} we present the SPRKD algorithm in full, including efficient Hessian eigenvalue computation via power iteration and Stochastic Lanczos Quadrature \citep{pearlmutter1994fast, ubaru2017stochastic, yao2020pyhessian, golmant2018hessianeigenthings}, Approximated Saddle Region (ASR) computation, Transfer Learning by Injection (TLI) for cross-architecture re-parameterization, an exponentially decaying Euclidean Distance Matrix transformation scheme for iterative student approaching, and Negative Hessian Eigensteps (NHE) and Perturbed Gradient Descent (PGD) for accelerated saddle escape \citep{jin2017escape}. Section~\ref{sec:experimental-setup} presents the experimental setup and Section~5 the empirical results, including top-line accuracies, Hessian eigenvalue spectral density (ESD) analysis, and loss landscape visualization, on remote-hospital malaria blood smear classification (a $4\times$ compression from a weak 25{,}546-parameter teacher), TinyImageNet (ResNet-101 to ResNet-18), and supplementary MNIST/CIFAR-100 benchmarks. Section~\ref{sec:discussion} discusses implications and broader context. Section~\ref{sec:conclusion} outlines limitations and next steps. All implementation code and reproduction scripts are publicly available on GitHub\footnote{\url{https://github.com/thetechdude124/SADDLE-POINT-RECRUITMENT-FOR-KNOWLEDGE-DISTILLATION}}.

\paragraph{Contributions.} (1) We re-frame KD from replication to curvature distillation, motivated by five complementary properties of saddle points in high-dimensional DNN loss landscapes \citep{dauphin2014identifying, zhang2021embedding, draxler2018barriers, liao2017theory, jastrzebski2019relation}. (2) We propose the SPRKD algorithm, a three-phase distillation pipeline that aggregates teacher saddle points, re-parameterizes them into the student via TLI, biases student parameters toward this ASR with an exponentially decaying Euclidean transformation, and accelerates descent via second-order NHE + PGD steps \citep{jin2017escape, alain2019negative}. (3) We demonstrate empirically that SPRKD removes the teacher-driven accuracy ceiling of traditional KD \citep{cho2019efficacy, stanton2021does}: on malaria classification, SPRKD outperforms RKD by 24.70 percentage points (94.80\% vs.~70.10\%) and matches Control-S to statistical equivalence despite distilling solely from a weak teacher (McNemar $p = 6.3 \times 10^{-87}$ vs.~RKD; $p = 1.0$ vs.~Control-S; test of \citealp{mcnemar1947}). (4) We characterize the optimization geometry of SPRKD students via Hessian ESDs, Hessian trace, gradient Lipschitz bounds, and 2-D loss landscape visualization \citep{ghorbani2019hessian, yao2020pyhessian, li2018visualizing, keskar2017large, hardt2016train}, finding that SPRKD students converge to substantially flatter, smoother, and more generalizable minima than both RKD and scratch-trained controls.

\section{Background: Why Saddle Points Matter for Knowledge Distillation}
\label{sec:five-tenets}

As DNN dimensionality climbs, the ratio of saddle points to local minima proliferates exponentially \citep{dauphin2014identifying}. High-loss regions of the optimization landscape are saturated with saddle points and are separated from dense clusters of local minima below \citep{dauphin2014identifying, goodfellow2015qualitatively}. In what follows we collect five complementary properties of saddle points that, when taken together, motivate the SPRKD methodology.

\subsection{Tenet 1: Saddle Point Proliferation in High Dimensions}
By the analysis of \citet{dauphin2014identifying}, the ratio of saddle points to local minima grows combinatorially with parameter-space dimensionality. The probability of local minima in higher-order spaces declines to infinitesimal amounts as implied by random-matrix and spin-glass analyses of DNN loss surfaces \citep{choromanska2015loss, dauphin2014identifying}, so the high-loss regions traversed by DNN optimizers consist overwhelmingly of saddles rather than minima. Successful optimization in deep models therefore depends strongly on the appropriate navigation of said saddle points (i.e. a good optimization technique must minimally ensure that optimization does not stall at degenerate saddles) \citep{dauphin2014identifying, jin2017escape}.

\subsection{Tenet 2: The Embedding Principle}
By the Embedding Principle of \citet{zhang2021embedding}, the loss landscape of a DNN of width $w+1$ contains all critical points of a DNN of width $w$. Local minima of the narrower network are mapped to degenerate saddle points (with strong likelihood) or to degenerate minima (with low likelihood) of the wider network, across all differentiable activations, loss functions, and data types. Saddle points encountered by a wide teacher therefore have a high probability of corresponding to convergence sites in a narrower student network.

\subsection{Tenet 3: Saddle Points as the Apex of Minimum-Energy Paths}
Work in molecular statistical mechanics and minimum-energy-path (MEP) algorithms finds that distinct local minima of a DNN loss surface are connected by low-loss paths whose error is close to that of the minima themselves \citep{draxler2018barriers, garipov2018loss}. These paths are bridged and upper-bounded by saddle points of marginally higher loss; the height of this loss barrier shrinks as dimensionality increases \citep{draxler2018barriers}. As a result, low-loss teacher saddle points lie on the apex of paths connecting strong local minima of the narrower student, and serve as natural waypoints for student re-exploration.

\subsection{Tenet 4: Basin-Fractal Decision Points}
Through Basin-Fractal theory \citep{liao2017theory}, saddle points serve as key decision points of raised loss separating basins of local minima and convergence sites at all levels of the loss landscape. Models on identical sides of peak-loss barriers yield equal or greater performance when averaged \citep{liao2017theory}. Saddles therefore function as natural \emph{routing} structures: distilling them propagates information about which basins are worth exploring, rather than about which specific minima are worth replicating.

\subsection{Tenet 5: Sharp Saddles, SGD Drift-Diffusion, and Untapped Descent}
Stochastic Gradient Descent (SGD) regimes exhibit a high-mean, low-variance \emph{drift} phase and a low-mean, high-variance \emph{diffusion} phase \citep{jastrzebski2019relation}. Across phase transitions, sharp points (large spectral radius) of strong further-descent potential are visited but rarely fully exploited: excessive SGD learning rate inhibits descent there, sending the optimizer to alternative regions and delaying convergence \citep{jastrzebski2019relation, alain2019negative}. These sharp saddles are intrinsically conducive to further descent (they form a local maximum along at least one Hessian eigendirection) \citep{dauphin2014identifying, alain2019negative} yet are left untapped by first-order optimizers \citep{jin2017escape}.

\paragraph{Synthesis.}
Saddle points encountered by teacher models are strongly likely to (a) map to potential convergence sites in lower-network (student) dimensions, (b) lie on the apex of paths connecting multiple low-loss basins, and (c) provide early and rapid descent to strong, generalizable solutions if negative-curvature directions are accurately identified and traversed. This synthesis is the foundation of SPRKD.

\section{Methodology: The SPRKD Algorithm}
\label{sec:method}

\paragraph{Fundamental Research Question.}
How might the selective and intelligent distillation of teacher saddle points, and by extension teacher optimization-landscape properties, reduce student size and minimize computational requirements while accelerating performance and generalization, catalyzing deep learning impact in critical low-compute settings without the need for expensive strong teachers?

\paragraph{Objectives.}
We seek to (i) generate strong, highly generalizable students from weak teachers within at most 3\% of strong teacher performance, eliminating the KD upper accuracy bound; (ii) employ identical loss functions across teachers and students during distillation, so that students learn on the actual task rather than directly replicating teacher logits; (iii) minimize total compute and development time across the distillation and training pipeline, eliminating simultaneous teacher inference during student training; and (iv) allow for distillation without accuracy or performance degradation, shifting KD from mimicry to a curvature-aware compression technique that does not depend on the availability of expensive, fully trained teachers.

\begin{figure*}[!htbp]
\centering
\includegraphics[width=\textwidth]{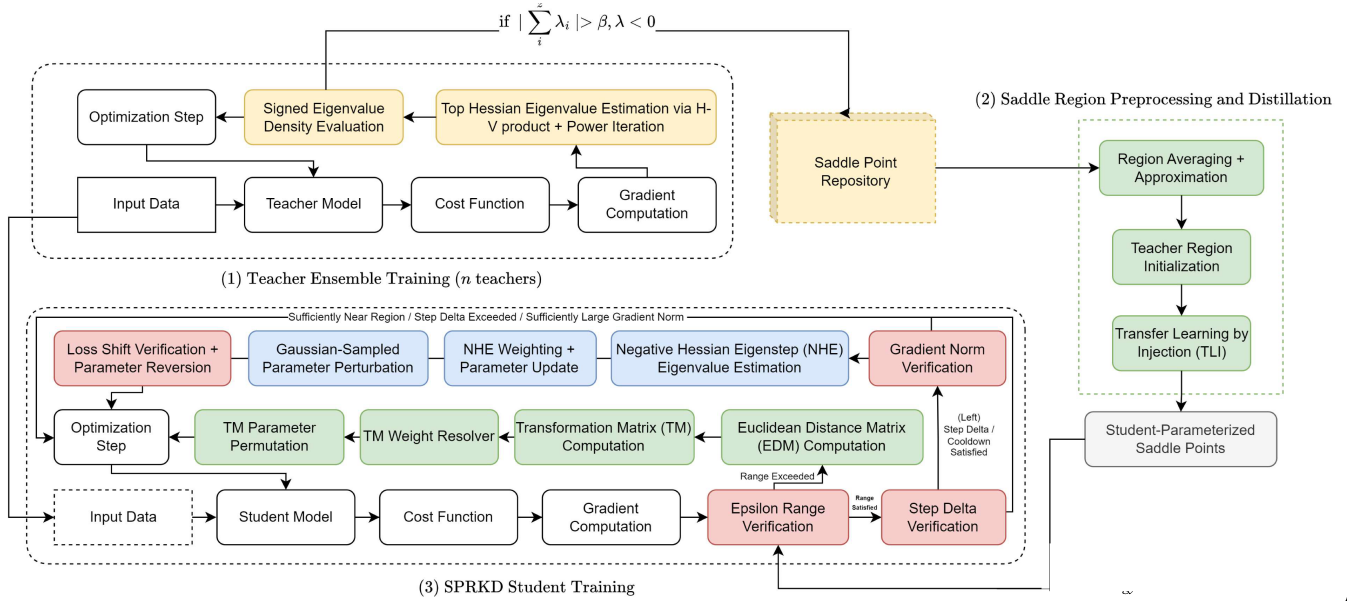}
\caption{SPRKD high-level algorithm overview and distillation methodology. \textbf{(1)} Teacher Ensemble Training tracks low-loss saddle points via signed eigenvalue density evaluation and top Hessian eigenvalue estimation, populating a Saddle Point Repository. \textbf{(2)} Saddle Region Preprocessing and Distillation averages and approximates the saddle region, then re-parameterizes it into the student via Transfer Learning by Injection (TLI). \textbf{(3)} SPRKD Student Training iteratively biases the student toward the ASR via a decaying Transformation Matrix, then escapes near-degenerate saddles via Negative Hessian Eigensteps (NHE) and Gaussian (Perturbed Gradient Descent) perturbations.}
\label{fig:overview}
\end{figure*}

\paragraph{Overview.}
SPRKD operates in three phases. Phase 1 trains a teacher ensemble while tracking low-loss saddle points via efficient Hessian eigenvalue estimation. Phase 2 aggregates the lowest-loss saddles across all teachers to form an Approximated Saddle Region (ASR), and re-parameterizes the ASR into student space via Transfer Learning by Injection (TLI). Phase 3 iteratively biases the student into the ASR via an exponentially decaying Euclidean Distance Matrix transformation, then applies Negative Hessian Eigensteps (NHE) and Gaussian (Perturbed Gradient Descent) perturbations to accelerate descent and escape near-degenerate saddles. Figure~\ref{fig:overview} summarizes the pipeline.

\subsection{Phase 1: Teacher Ensemble Training and Saddle Tracking}

We train an ensemble of teacher models on the true data distribution. Each teacher may be a weak model trained for only a few epochs; aggregating saddle points across \emph{multiple} weak teachers yields a richer Approximated Saddle Region than a single teacher provides, because distinct weak teachers encounter distinct saddle structures during their brief training trajectories. In our experiments, multi-teacher saddle aggregation consistently outperformed single-teacher distillation, consistent with prior ensemble-distillation findings \citep{asif2020ensemble}. In conjunction with teacher training, we evaluate the top $z$ Hessian eigenvalues via power iteration every $k$ steps to detect saddle points (we initially employed the Lanczos method via \texttt{hessian-eigenthings} \citep{golmant2018hessianeigenthings} but switched to power iteration after accuracy concerns in early trials). Computational cost is negligible at approximately one backpropagation per monitoring step, since Hessian-vector-product-based estimation finds properties of the Hessian, such as eigenvalues, without explicitly storing the full $H \in \mathbb{R}^{n \times n}$ matrix \citep{pearlmutter1994fast, coleman1998hessian, ghorbani2019hessian}. In practice we use $z = 4$ for student models and estimate the top 2 to 20 eigenvalues for teacher models; $k \in [1, 50]$. The Hessian trace is computed via the Hutchinson estimator \citep{hutchinson1990stochastic, ghorbani2019hessian, yao2020pyhessian} with a Rademacher probe vector. Eigenvalue Spectral Density is computed via Stochastic Lanczos Quadrature \citep{ubaru2017stochastic, ghorbani2019hessian, yao2020pyhessian}, applying a Gaussian kernel to an initial eigenvalue estimate followed by Gaussian quadrature, because the full ESD problem is otherwise intractable at DNN scale \citep{ghorbani2019hessian}.

\paragraph{Identifying Strong Saddle Points.}
SGD optimizers encounter nontrivially sharp points during phase transitions at which convergence fails, inhibiting performance. Such points allow for rapid early descent to low-loss minima clusters upon student distillation, provided accurate traversal, and are therefore strongly beneficial to student models \citep{jastrzebski2019relation}. Points of strongest potential for further descent are identified as follows. Let $\Lambda$ denote the set of all Hessian eigenvalues at a given step. The parameter state is taken as a strong (sharp) saddle point if the negative eigenvalue density and the magnitude of leading negative eigenvalues are sufficient:
\begin{equation}
\Big| \sum_{\lambda_i < -\beta} \lambda_i \Big| > \beta,
\quad \sum_{i} \lambda_i^{+} > \alpha \sum_{i} |\lambda_i^{-}|,
\end{equation}
with hyperparameters $\alpha \approx 0.4$ and $\beta = 7$ tuned on a small validation sweep. Parameter snapshots at all qualifying points are stored in a saddle point repository for later aggregation. Crucially, this is a passive monitoring step: it adds no additional inference cost beyond the per-$k$-step Hessian-vector-product call.

\subsection{Phase 2: Saddle Region Approximation and Distillation}

\paragraph{Approximated Saddle Region (ASR).}
The lowest-loss (degenerate) saddle points across Phase 1 are averaged to calculate an Approximated Saddle Region with strong likelihood of mapping to minima or minima-populated subspaces as network width diminishes (Tenet 2). The ASR aggregates the sharpest, highest-potential-descent points across the SGD drift / diffusion transition (Tenet 5), and is treated as the target for student initialization rather than as a final convergence site.

\paragraph{Transfer Learning by Injection (TLI).}
Distinct DNN parameter spaces yield dissimilar optimization landscapes; the ASR therefore must be re-parameterized in terms of the student \citep{czyzewski2021weights}. Neural Architecture Search and Fast Network Adaptation are insufficient for the complex, mismatched architectures encountered in practice. We use TLI \citep{czyzewski2021weights}: (i) group student and teacher layers by traversing the PyTorch autograd graph operations \texttt{AddBackward0}, \texttt{MulBackward0}, and \texttt{CatBackward}; (ii) iteratively modify the student computational graph until it becomes structurally similar to the teacher's; (iii) inject convergent parameters into the student via center crop and resize operations. TLI reparameterizes the ASR into student space and provides a stronger initialization than random weights, though SPRKD does not directly initialize the student at the injected ASR (see below). The embedding principle (Tenet 2) requires matched computational depth and a strictly narrower student than teacher. This is a substantive deployment constraint: deep networks such as ResNets are not shallow in the embedding-principle sense, and successful cross-architecture distillation still depends on explicit layer pairing under TLI rather than on an automatic depth reduction.

\paragraph{Iterative Approaching, not Direct Initialization.}
An important design choice is that we do \emph{not} directly initialize the student at the ASR. Direct initialization would risk converging onto irregular or deformed saddle points distilled despite negative eigenvalue density evaluation. Instead, we iteratively approach the ASR within a user-specified $\epsilon$-delta, after which standard first-order optimization is free to anneal the most promising descent paths. This yields diverse saddle exploration as a product of unique student initializations.

\subsection{Phase 3: Student Saddle Targeting and Acceleration}

\subsubsection{Iterative ASR Approaching via Decaying Transformation Matrices.}
Let $S_i$ and $T_i$ denote the $i$-th student parameter and saddle point matrices, respectively. The transformation matrix is given by $M_i = T_i \oslash S_i$ (elementwise division) and is applied to the student as
\begin{equation}
S_i' = S_i \odot (l \times M_i),
\end{equation}
where $l = -\,2^{-t/10} + 1$ is an exponentially decaying weight, asymptotically approaching $1$ as the transformation matrix shrinks (similar to ADAM's decaying moment estimates \citep{kingma2015adam}). Iteration continues until the Euclidean distance between current and target parameters is within $\epsilon$:
\begin{equation}
\epsilon \;\ge\; \max_{rc,\, r=c} \sqrt{\lambda_{\max}\big((S_i' - T_i)^\top (S_i' - T_i)\big)}.
\end{equation}
We use $\epsilon = 0.1$ in all experiments. This procedure runs for 10 to 200 steps depending on model depth and layer size.

\subsubsection{Negative Hessian Eigensteps and Gaussian Perturbations.}
Iterative ASR approaching makes the student more likely to land on near-degenerate saddles, which SGD-based optimizers may take exponential time to escape \citep{dauphin2014identifying, jin2017escape}. We therefore augment standard student training with two ingredients after Transformation Matrix completion:

\begin{enumerate}
\item \textbf{Saddle stagnation monitoring.} At every step, we monitor the gradient L2-norm. If $\|\nabla \mathcal{L}\|_2 \leq j$ for $j = 0.02$, the student is flagged as stagnating.
\item \textbf{Largest-negative eigenvalue extraction and NHE.} We compute the largest-magnitude negative Hessian eigenvalue $\lambda$ and corresponding eigenvector $v$ via power iteration on Hessian-vector products \citep{pearlmutter1994fast, golmant2018hessianeigenthings}, then take a step inversely proportional to $|\lambda|$ along the negative-curvature direction:
\begin{equation}
\theta_{t+1} = \theta_t - \frac{1}{|\lambda|}\, \big(\nabla \mathcal{L} \odot v \odot v\big),
\end{equation}
where $\odot$ denotes element-wise product on each parameter tensor and $v$ is the Hessian eigenvector block aligned with that tensor (our released implementation matches this form; earlier slide decks used equivalent scalar shorthand). Descent closely matches the curvature of the loss landscape and is more accurate than first-order steps alone \citep{alain2019negative}. SGD often bounces around sharp saddle regions due to excessive learning rate \citep{jastrzebski2019relation}; NHE prevents this, yielding smoother descent. SPRKD students exhibit fewer residual negative eigenvalues at convergence because negative curvature regions are more completely traversed during training \citep{alain2019negative}.
\item \textbf{Gaussian (PGD) Perturbation.} We then apply a Gaussian-sampled perturbation $\theta_{t+1} = \theta_t - \xi$, $\xi \sim \mathcal{N}(0, 0.1)$, moving the optimizer to a higher-magnitude region of the gradient field \citep{jin2017escape}.
\item \textbf{Verification and Revert.} If NHE + perturbation jointly fail to decrease loss, we revert to the pre-NHE checkpoint. This guards against catastrophic perturbation in irregular curvature regions.
\end{enumerate}

\paragraph{Benefits and Convergence Properties.}
NHE explicit curvature identification and traversal hastens saddle-point escape and accelerates descent to minima-populated subspaces below \citep{alain2019negative}. The ASR is comprised of the sharpest SGD-phase saddle points (strongly degenerate in teacher dimensions); NHE enables exact and precise descent of high-potential distilled second-order information, maximizing teacher descent utility otherwise ignored by first-order approaches. Gaussian perturbations move the student to high-magnitude portions of the gradient field upon stagnation, allowing for faster escape times \citep{jin2017escape}. Cumulatively, SPRKD is designed to escape near-degenerate saddles encountered after ASR approaching. We view the PGD component as consistent with the escape-time analysis of \citet{jin2017escape}, but we do \emph{not} claim that the full ASR + NHE + PGD pipeline inherits the same second-order stationary-point convergence guarantee as standalone PGD; a unified proof for the combined optimizer remains future work (Section~\ref{sec:conclusion}).

\paragraph{Computational and Time Complexity.}
We do not report wall-clock epoch tables in this work; the efficiency argument is structural rather than a timing benchmark. Replication-based KD typically requires (i) training a \emph{strong} teacher to high task accuracy before distillation begins, and (ii) simultaneous teacher and student forward passes on every student minibatch \citep{stanton2021does, gou2021kd}, coupling student optimization to persistent teacher inference cost. SPRKD targets the opposite regime: distillation from \emph{weak} teachers trained for only a few epochs (2 epochs in Experiment~1), with saddle monitoring amortized to one Hessian-vector-product call every $k$ steps during that short Phase~1. After the ASR is built and injected, Phase~3 trains the student on task labels alone---no live teacher in the loop---so each standard optimization step remains first-order SGD with per-iteration complexity $\mathcal{O}(n)$ in the student parameter count $n$, the same asymptotic order as training Control-S from scratch. Intermittent NHE and PGD add occasional Hessian-vector products when stagnation is detected, but they do not reintroduce the standing $\mathcal{O}(n_{\text{teacher}})$ teacher forward pass that dominates replication-based KD. The dominant compute savings are therefore \emph{upstream}: strong-teacher pretraining is not a prerequisite for obtaining a high-performing compressed student.

\section{Experimental Setup}
\label{sec:experimental-setup}

We design two experiments to test whether SPRKD methodologies yield students of significantly reduced size and faster generalization from weak teachers, with reduced training time and compute compared to standard KD and scratch-trained controls. All experiments were implemented in PyTorch \citep{paszke2019pytorch} and conducted on a single Google Colab Tesla GPU and a laptop with an Intel i3 processor; no multi-GPU cluster was required. The full project spanned approximately seven months of development and experimentation. Hessian trace, ESD, and top-eigenvalue analyses in Section~\ref{sec:results} use \texttt{PyHessian} \citep{yao2020pyhessian}; online saddle monitoring during teacher training uses \texttt{hessian-eigenthings} \citep{golmant2018hessianeigenthings}.

\subsection{Experiment 1: Malaria Blood Smear Classification}

\paragraph{Motivation.}
Adjacent work aims to distill large Convolutional Neural Networks (CNNs) into lightweight form factors via Response Knowledge Distillation (RKD) for smartphone-based malaria diagnosis in remote, low-cost laboratories in developing nations where equipment availability is restricted \citep{fuhad2020malaria}. We leverage SPRKD to distill a weak variant of a 25{,}546-parameter CNN (drawn from \citet{fuhad2020malaria}) into a 6{,}430-parameter student, comparing performance and optimization metrics to a strong 25{,}546-parameter teacher and to its million-parameter parent.

\paragraph{Dataset and Preprocessing.}
We use the public Malaria blood smear dataset from the U.S.~National Library of Medicine, consisting of 27{,}558 annotated cell images with a 50/50 class split between parasite-infected and healthy cells \citep{nlmmalaria}. Preprocessing follows \citet{fuhad2020malaria}: 32$\times$32 image size, batch size 64, 0.75 train/validation split, no data augmentation. All student and control models train for 500 epochs at 323 gradient steps per epoch; validation metrics are logged at 175 evenly spaced checkpoints (Figure~\ref{fig:loss-acc}).

\begin{figure}[t]
\centering
\includegraphics[width=\columnwidth]{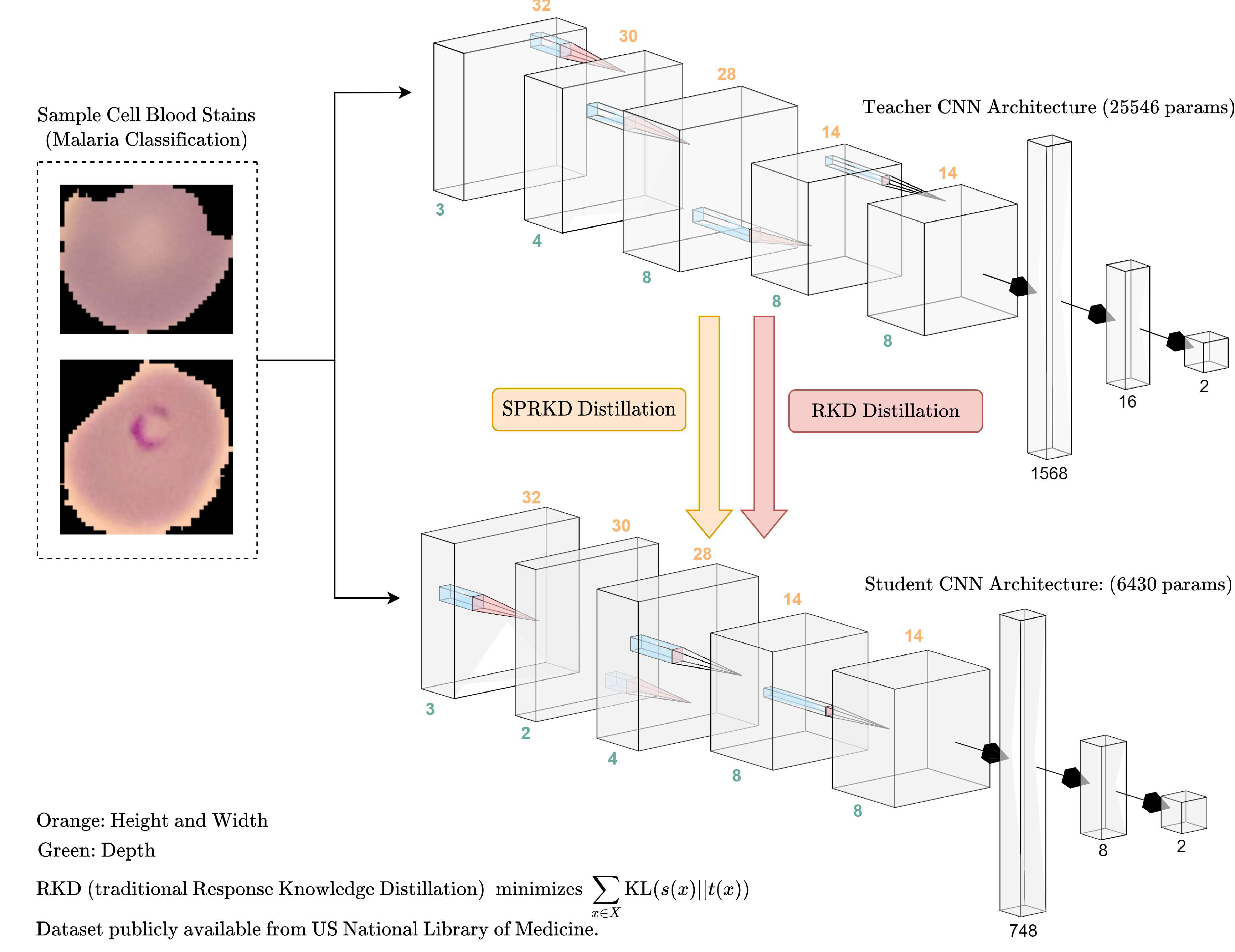}
\caption{Experiment 1 distillation setup. A 25{,}546-parameter teacher CNN is distilled into a 6{,}430-parameter student CNN (a $4\times$ compression) on malaria-infected blood smear images. Cube widths denote spatial resolution (orange) and channel depth (green); the rightmost columns indicate flattened dense-layer sizes (1{,}568 for the teacher, 748 for the student). SPRKD distillation uses second-order saddle-region information from the weak teacher; RKD uses the standard logit-matching loss $\mathcal{L}_{KD} = \sum_{x \in X} \mathrm{KL}(s(x) \| t(x))$ for direct comparison. Dataset: U.S.~National Library of Medicine \citep{nlmmalaria}.}
\label{fig:malaria-arch}
\end{figure}

\paragraph{Models.}
We train four cumulative models: a Strong Control Teacher (Control-T) trained from scratch on the optimization task, a Weak Teacher (employed in distillation; trained for 2 epochs), a Control Student (Control-S, trained from scratch with the same architecture as the SPRKD student), and an RKD student (distilled from the Weak Teacher using standard logit-matching \citep{hinton2015distilling}). The SPRKD student is distilled from the Weak Teacher only. We conduct 5 trials per experiment and average all performance metrics (top-1). Figure~\ref{fig:malaria-arch} summarizes the teacher / student architectures and their respective parameter counts.

\paragraph{Hypotheses.}
We hypothesize that SPRKD will exceed RKD by substantial margins and match Control-S and Control-T performance to statistical equivalence; that RKD will replicate Weak Teacher performance (near \mbox{70\%} validation accuracy); and that SPRKD will advance beyond the weak-teacher ceiling.

\subsection{Experiment 2: Residual Networks on TinyImageNet}

\paragraph{Motivation.}
A natural question is whether SPRKD scales to deeper, industry-standard architectures and richer data. We execute SPRKD distillation between ResNet-101 (43 M params) and ResNet-18 (11 M params) \citep{he2016resnet} on the TinyImageNet dataset, a subset of the complete ImageNet challenge involving 200 classes and 100k samples \citep{le2015tinyimagenet, russakovsky2015imagenet}.

\paragraph{Setup.}
Preprocessing: 64$\times$64 image size, 500 images per class across 200 classes, 2 to 10 epochs (2{,}813 steps/epoch), 0.75 train/validation split. Experimental methodology is kept consistent with Experiment 1 to enable a direct comparison to existing literature on ResNet-18 TinyImageNet capacity.

\subsection{Supplementary Benchmarks: MNIST and CIFAR-100}

To assess generalization beyond the malaria task, we additionally evaluate SPRKD on MNIST (MLP architectures) \citep{lecun1998gradient} and CIFAR-100 (CNN architectures) \citep{krizhevsky2009learning} under the same weak-teacher SPRKD protocol.

\begin{figure*}[!t]
\centering
\includegraphics[width=\textwidth]{figures/fig_loss_acc.png}
\caption{Validation loss (left) and accuracy (right) at 175 logged checkpoints over 500 training epochs for SPRKD, Control-S, and RKD on malaria blood smear classification (Experiment 1). The horizontal axis indexes validation checkpoints (not raw epoch count). SPRKD's descent is smoother (less noisy) than Control-S and overtakes it at roughly 10 checkpoints; the early slowdown reflects TM application as the ASR is approached, after which descent rapidly steepens. RKD distillation loss is near-zero (perfect teacher replication), yet task accuracy plateaus near \mbox{70\%} (Table~\ref{tab:malaria}), failing to surpass the weak teacher and generalize. SPRKD converges to \mbox{94.80\%} validation accuracy, overcoming the replication-induced accuracy bound.}
\label{fig:loss-acc}
\end{figure*}

\section{Results}
\label{sec:results}

\subsection{Top-Line Accuracy: Experiment 1}

Table~\ref{tab:malaria} reports CNN model performance on the malaria task, averaged across 5 trials.

\begin{center}
\small
\setlength{\tabcolsep}{3pt}
\begin{tabular}{lrrrrr}
\toprule
\textbf{Model} & \textbf{Ep.} & \textbf{Params} & \textbf{Val.~Loss} & \textbf{Acc.} & \textbf{Err.} \\
\midrule
Control-Teacher & 500 & 25{,}546 & 0.364 & \mbox{94.50} & \mbox{5.50} \\
Teacher (Weak)  & 2   & 25{,}546 & 0.583 & \mbox{70.13} & \mbox{29.87} \\
Control-Student & 500 & 6{,}430  & 0.364 & \mbox{94.47} & \mbox{5.53} \\
RKD             & 500 & 6{,}430  & $\!\sim\!0$ & \mbox{70.10} & \mbox{29.90} \\
\textbf{SPRKD}  & 500 & 6{,}430  & \textbf{0.361} & \textbf{\mbox{94.80}} & \textbf{\mbox{5.20}} \\
\bottomrule
\end{tabular}
\captionof{table}{CNN model performance on remote malaria blood smear classification (averaged across 5 trials). RKD validation loss is the distillation loss between the RKD student and the Weak Teacher (near zero: $4.90\!\times\!10^{-6}$, indicating near-perfect logit replication); all other models employ cross-entropy loss between true and predicted labels.\label{tab:malaria}}
\end{center}

Per-checkpoint validation loss and accuracy curves are shown in Figure~\ref{fig:loss-acc}.

SPRKD outperforms RKD by 24.70 percentage points (\mbox{94.80\%} vs.~\mbox{70.10\%}) and matches Control-T and Control-S performance to statistical equivalence (\mbox{94.80\%} vs.~\mbox{94.50\%} and \mbox{94.47\%}; McNemar $p = 1.0$ vs.~Control-S). It does so while distilling solely from the Weak Teacher (\mbox{70.13\%} accuracy at 2 epochs), with 4$\times$ fewer parameters than the Strong Teacher, and without simultaneous teacher inference during training \citep{stanton2021does}. We assess paired validation predictions with McNemar's test \citep{mcnemar1947}, which yields a statistically significant difference between SPRKD and RKD ($p = 6.3 \times 10^{-87}$). RKD distillation loss is near zero ($4.90 \times 10^{-6}$), confirming near-perfect teacher logit replication \citep{hinton2015distilling}, yet task performance remains near the weak-teacher ceiling (\mbox{70.10\%}), validating the central critique of replication-based KD \citep{cho2019efficacy, stanton2021does, yuan2020revisiting}.

\subsection{Top-Line Accuracy: Experiment 2 and Supplementary Benchmarks}

Experiment~2 (ResNet on TinyImageNet) and the MNIST/CIFAR-100 supplementary benchmarks are reported as \emph{preliminary} qualitative findings (no dedicated results table in this submission); Table~\ref{tab:malaria} contains the fully tabulated head-to-head comparison. Across these supplementary runs, SPRKD shows the same qualitative pattern as Experiment~1: higher task accuracy than response KD and smoother optimization than control baselines under weak-teacher distillation. On CIFAR-100, both SPRKD and scratch-trained controls are evaluated at epoch~10 under the same weak-teacher protocol: SPRKD reaches \mbox{96\%} validation accuracy versus \mbox{88\%} for the control (an 8 percentage-point advantage at the matched epoch~10 checkpoint). On MNIST, at epoch~10, SPRKD reaches \mbox{99\%} versus \mbox{97\%} for the control under the same weak-teacher protocol. On malaria, SPRKD exceeds RKD by 24.70 percentage points (Table~\ref{tab:malaria}). All supplementary runs use weak teachers trained for a fraction of conventional KD teacher time (approximately 1\% in Experiment~1).

\subsection{Hessian Eigenvalue Spectral Density (ESD) Analysis}

\begin{figure*}[!t]
\centering
\includegraphics[height=0.32\textheight,keepaspectratio]{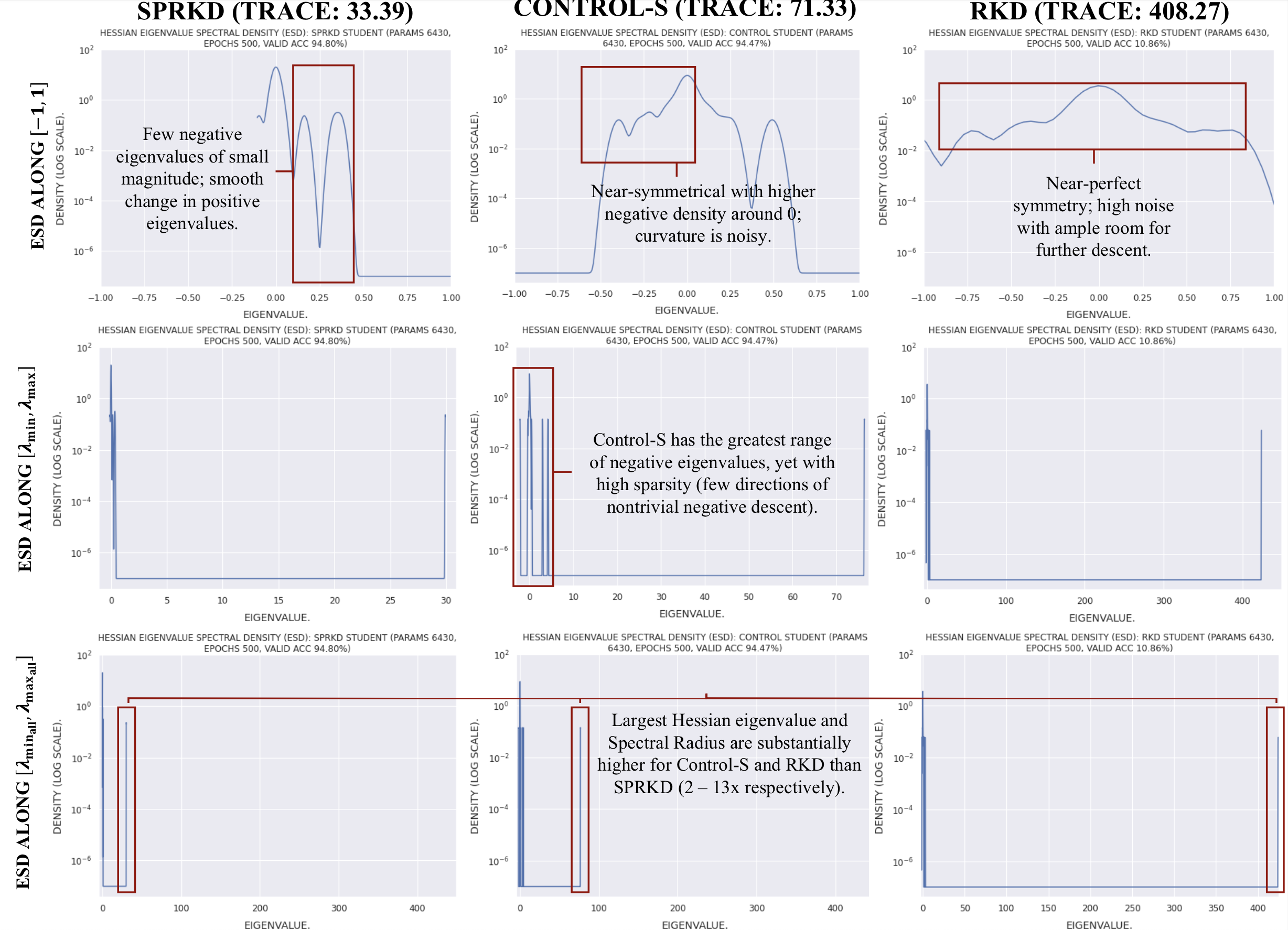}
\caption{Hessian Eigenvalue Spectral Density (ESD) plots across SPRKD, Control-S, and RKD students upon convergence (epoch 500), computed with \texttt{PyHessian} \citep{yao2020pyhessian}. Column headers report Hutchinson-estimated Hessian trace \citep{hutchinson1990stochastic, yao2020pyhessian}; panel subtitles report validation accuracy at the ESD measurement checkpoint. Each row visualizes the ESD over a different eigenvalue range: $[-1, 1]$ (top), $[\lambda_{\min}, \lambda_{\max}]$ (middle), and $[\lambda_{\min,\text{all}}, \lambda_{\max,\text{all}}]$ (bottom). SPRKD (trace 33.39, \mbox{94.80\%} accuracy) has few negative eigenvalues of small magnitude and a smooth change in positive eigenvalues. Control-S (trace 71.33, \mbox{94.47\%} accuracy) is near-symmetric with higher negative density around zero and noisier curvature. RKD (trace 408.27) exhibits near-perfect symmetry, high noise, and ample room for further descent neglected by replication-based distillation. The largest Hessian eigenvalue and spectral radius are substantially higher for Control-S and RKD than for SPRKD (2 to 13$\times$).}
\label{fig:esd}
\end{figure*}

The Hessian matrix $H_f(\theta)$ encodes second-order curvature of the loss landscape at parameters $\theta$ \citep{ghorbani2019hessian, goodfellow2015qualitatively}. Its eigenvalues quantify how strongly the loss changes along corresponding eigenvector directions: negative eigenvalues indicate saddle structure (descent remains possible along at least one direction), while the magnitude of the largest eigenvalue (spectral radius) and the trace (sum of eigenvalues) characterize sharpness and overall curvature \citep{ghorbani2019hessian, keskar2017large}. We measure the Hessian Eigenvalue Spectral Density (ESD) of all three students at epoch 500, alongside the Hessian trace. ESDs are estimated via Stochastic Lanczos Quadrature \citep{ubaru2017stochastic, ghorbani2019hessian, yao2020pyhessian} without explicitly materializing $H_f \in \mathbb{R}^{n \times n}$, using the \texttt{PyHessian} implementation \citep{yao2020pyhessian}; the trace is computed via the Hutchinson estimator \citep{hutchinson1990stochastic, yao2020pyhessian} with a Rademacher probe vector. Figure~\ref{fig:esd} visualizes the resulting ESDs over three eigenvalue ranges.

\begin{itemize}
\item \textbf{SPRKD (trace 33.39).} Low negative eigenvalue density and the smallest trace and spectral radius. Few negative eigenvalues of small magnitude; smooth change in positive eigenvalues. This implies near-perfect convergence to highly degenerate (flat) minima \citep{keskar2017large, chaudhari2019entropy}.
\item \textbf{Control-S (trace 71.33).} ESD is near symmetric around 0, with double the trace and spectral radius of SPRKD. Minima convergence indicates greater imperfection with strong room for further descent. Large trace further implies sharper minima convergence \citep{keskar2017large}; curvature is noisy.
\item \textbf{RKD (trace 408.27, \mbox{70.10\%} task accuracy).} Largest trace and spectral radius, with substantial room for further descent that is neglected by the distillation loss. The RKD panel subtitle shows \mbox{10.86\%}; epoch-500 task accuracy is \mbox{70.10\%} (Table~\ref{tab:malaria}). Convergence is improper with a high degree of instability around highly sharp minima; near-symmetric ESD with high noise.
\end{itemize}

Assuming the loss function is $h$-gradient-Lipschitz, the largest Hessian eigenvalues are bounded by $h$ \citep{hardt2016train}. Models with small $\lambda_{\max}$ can therefore be expected to exhibit more stable gradients and descent. SPRKD's substantially smaller $\lambda_{\max}$ and trace, combined with the loss landscape observations below, are consistent with a smaller effective gradient Lipschitz constant.

\subsection{Loss Landscape Visualization}

We project the loss landscape near convergence onto the top two Hessian eigenvectors (perturbed along $[\lambda_{\min}, \lambda_{\max}]$), following standard filter-normalization and eigenvector-projection practice \citep{li2018visualizing, horoi2022exploring, yao2020pyhessian}; Figure~\ref{fig:landscape} shows the resulting 3-D surfaces and 2-D heatmaps:

\begin{figure*}[!t]
\centering
\includegraphics[height=0.36\textheight,keepaspectratio]{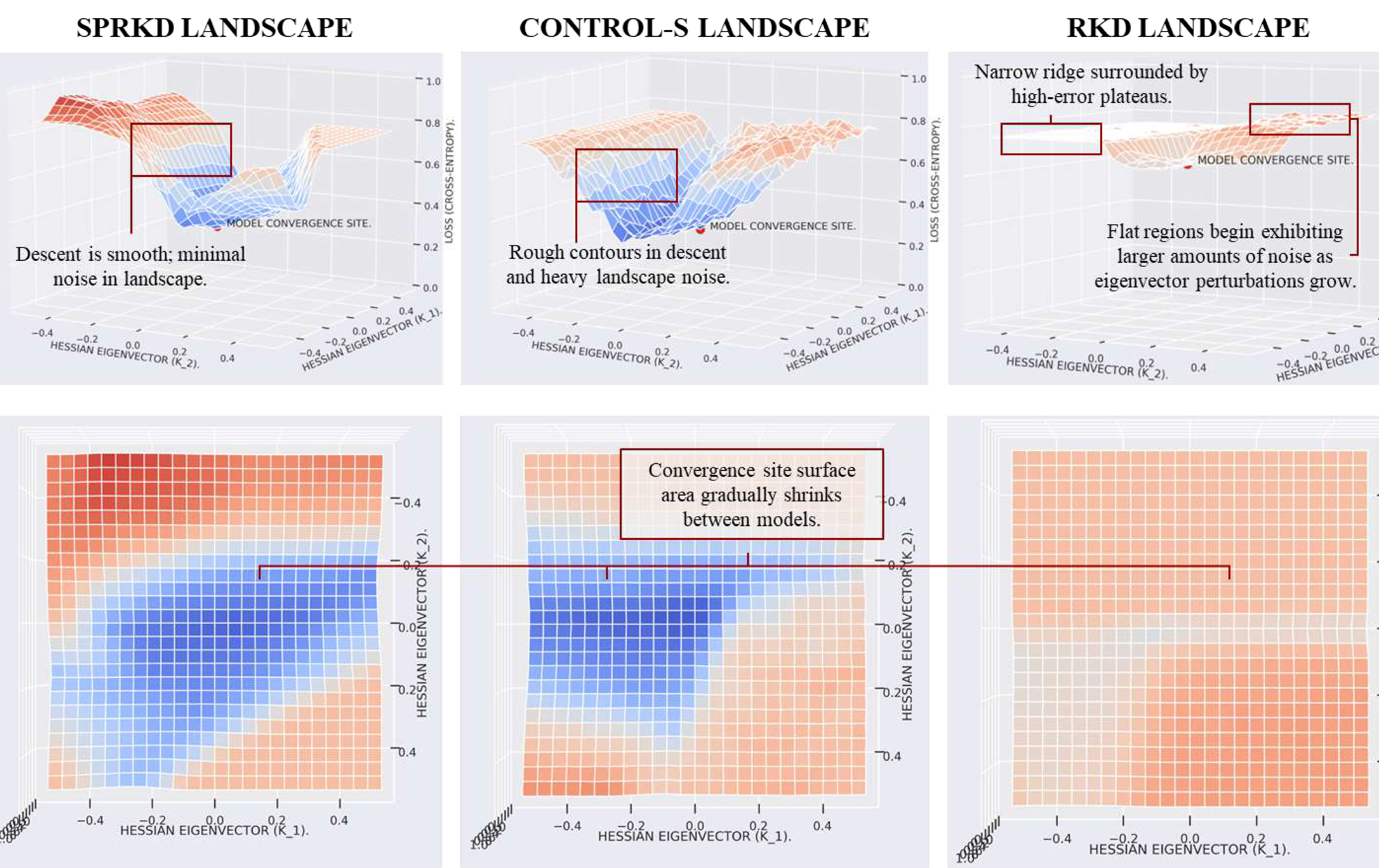}
\caption{Loss landscapes at convergence (epoch 500), perturbed across the top two Hessian eigenvectors. Each column corresponds to one student (SPRKD, Control-S, RKD); the top row shows the 3-D loss surface (cross-entropy on the $z$-axis) and the bottom row shows a top-down 2-D heatmap of the same surface. The model convergence site is annotated on each 3-D plot. SPRKD descends smoothly with minimal landscape noise; Control-S exhibits rougher contours and heavier landscape noise; RKD converges on a narrow ridge surrounded by high-error plateaus, with flat regions exhibiting larger amounts of noise as eigenvector perturbations grow. The convergence-site surface area shrinks substantially between SPRKD and the two baselines, consistent with SPRKD reaching wider minima.}
\label{fig:landscape}
\end{figure*}

\begin{itemize}
\item \textbf{SPRKD landscape.} Wider surface area for low-loss regions (resembling wide minima), with smooth descent pathway and minimal noise \citep{keskar2017large, chaudhari2019entropy}. Descent is smooth; landscape noise is minimal.
\item \textbf{Control-S landscape.} Sharper and smaller convergence site (coinciding with Hessian trace analysis). Higher noise, random contours, and sharp descent regions at the end of training.
\item \textbf{RKD landscape.} Convergence on a ridge between two high-loss plateaus, of high steepness and small surface area. A firmly sharp site surrounded by degenerate low-performance regions.
\end{itemize}

\section{Discussion}
\label{sec:discussion}

\paragraph{SPRKD vs.~Training a Control Student from Scratch.}
A natural question is why one should use SPRKD instead of simply training a Control Student on task labels. The core objective of both KD and SPRKD is not merely a compressed model, but a compressed model that preserves the generalization characteristics of a larger teacher while \emph{exceeding} teacher accuracy. Replication-based KD fails this objective: the student is upper-bounded by teacher performance and acts as label-smoothing regularization rather than substantive knowledge transfer \citep{cho2019efficacy, stanton2021does, yuan2020revisiting}. Training from scratch on labels alone ignores the second-order landscape information that teachers accumulate during optimization \citep{ghorbani2019hessian, alain2019negative}. SPRKD distills this information via saddle regions, enabling the student to converge faster, more stably, and to wider minima than either RKD or scratch training, as confirmed by our ESD and loss landscape analyses.

\paragraph{Removal of the Teacher-Driven Accuracy Bound.}
SPRKD approaches appear to remove the teacher-induced upper bound on student accuracy after a brief ASR-iterative-approaching-induced delay. Models exhibit faster, more stable, and farther descent while maximizing teacher domain knowledge (the distilled ASR boosts convergence speed). The RKD student, despite having substantial potential for further descent (high negative eigenvalue density), plateaus at almost identical accuracy as the Weak Teacher, confirming both theoretical and empirical observations on KD limitations \citep{cho2019efficacy, stanton2021does, yuan2020revisiting}.

\paragraph{Stronger Convergence and Smaller Gradient Lipschitz Constants.}
SPRKD models exhibit lower Hessian trace, smoother loss landscapes, and smaller highest eigenvalues (Figures~\ref{fig:esd} and~\ref{fig:landscape}). Hessian trace and largest eigenvalue typically decrease as convergence to stationary points occurs (a sign of training progression \citep{alain2019negative}). Combined with low negative eigenvalue density, this implies full employment of negative curvature regions to maximize descent. A smaller $\lambda_{\max}$ further implies a smaller gradient Lipschitz constant, a potent measure of landscape and model stability \citep{hardt2016train}. Combined with qualitative observations, these findings suggest SPRKD methodologies remove noise from optimization via accurate, teacher-vetted negative saddle region traversal, positing higher stability of optimization sites.

\paragraph{Wider Minima and Generalization.}
SPRKD students converge to wider minima than both Control-S and RKD models (Figure~\ref{fig:landscape}), associated extensively with stronger generalization, robustness to dataset and activation noise, and clustering below saddle points \citep{chaudhari2019entropy, keskar2017large, li2018visualizing}. This implies convergence to minima stronger than both RKD methods and even Control-S models. While the Control model also achieves relatively wide minima (to a lesser extent than SPRKD), RKD fails on this dimension and accordingly exhibits the lowest task performance.

\paragraph{Sources of Error and Considerations.}
We considered the top 20 and top 2 eigenvalues for teacher and student models, respectively. While these quantities are sufficient for accurate saddle detection in the regimes we tested, including larger samples may further improve ASR construction and NHE accuracy. The learning capacity of the underlying architecture also plays a role in maximum achievable accuracy: the difference between Control-S and SPRKD may grow for more complex tasks where Control-S itself struggles to converge well.

\paragraph{Implications for KD.}
Through leveraging teachers as information oracles regarding task optimization landscapes, SPRKD helps shift KD from replication to teacher-driven generalization and performance improvement \citep{gou2021kd, alkhulaifi2021kd}. Rather than students replicating teachers, the teacher provides critical second-order knowledge to students that allows for boosted task proficiency, accessible regardless of teacher performance \citep{ghorbani2019hessian, yao2020pyhessian}.

\paragraph{Generalization Beyond the Malaria Task.}
SPRKD is not specific to the loss landscape of malaria classification. As long as the theoretical properties in Section~\ref{sec:five-tenets} hold, the methodology should apply to any stochastic DNN loss with saddle points and minima. All DNNs exhibit such structure in high dimensions \citep{dauphin2014identifying, choromanska2015loss}. The primary architectural constraint is that the student must share depth with but be narrower than the teacher to permit the embedding principle (Tenet 2) \citep{zhang2021embedding}. MNIST, CIFAR-100, and TinyImageNet results provide preliminary evidence that SPRKD transfers across datasets and architectures; broader validation on transformers, non-image modalities, and production edge deployments remains ongoing work.

\paragraph{Edge Deployment and Why Cloud-Only ML Is Insufficient.}
Most contemporary ML is cloud-based, where model size is less constrained. SPRKD matters because mission-critical applications cannot tolerate cloud latency: too much data must be processed too quickly (for example, ICU monitoring, autonomous navigation, real-time fault detection in energy infrastructure) \citep{alkhulaifi2021kd}. Radios consume substantial energy and internet access is not guaranteed on embedded devices in hospitals, traffic systems, and remote industrial sensors. Privacy further motivates local inference: hospital ECG and diagnostic data should not need to leave the device. SPRKD enables deployment of high-performance models on connected edge and IoT devices projected worldwide \citep{ericsson2022iot, alkhulaifi2021kd}, from hospital microscopes for automated bacteria classification (where microbiologist shortages motivate automated screening) to wind-turbine orientation controllers that adjust based on local sensor data. Reported edge deployments in industrial settings have documented substantial reductions in operational cost and data transfer relative to cloud-only inference \citep{alkhulaifi2021kd}. The same logic extends to distilling large language models: transformer architectures with hundreds of billions of parameters require thousands of cloud servers \citep{strubell2019energy}; SPRKD-style landscape-aware distillation is one path toward deploying capable models on smartphones and embedded hardware for critical, latency-sensitive applications.

\paragraph{Multi-Teacher Weak Ensembles.}
Training multiple weak teacher models and aggregating their saddle points into a single ASR consistently improved student performance relative to single-teacher distillation in our experiments \citep{asif2020ensemble}. Each weak teacher, trained for only a few epochs, encounters different saddle structures; averaging the lowest-loss saddles across the ensemble yields a denser, more informative target region for the student. This makes SPRKD particularly attractive in data-scarce settings where no single strong teacher is available but several cheap weak teachers can be trained independently.

\section{Conclusion, Limitations, and Future Work}
\label{sec:conclusion}

In the final analysis, all experimental results satisfy our engineering objectives and confirm the end hypothesis, suggesting that SPRKD: (i) converges to more generalizable and higher-quality sites; (ii) enables faster time-to-convergence and descent; (iii) smoothens the optimization landscape; (iv) removes replication-induced accuracy barriers in distillation performance; (v) maximizes student learning capacity via embedded teacher domain and curvature knowledge irrespective of teacher strength; and (vi) enables strong student distillation from computationally cheap weak teachers, broadening edge applications in significantly less compute and development time.

\paragraph{Key Applications.}
SPRKD has direct implications for compressing large DNNs for edge deployment without strong teacher training; enabling high-powered models in data-scarce and low-latency settings (hospitals, autonomous vehicles, energy infrastructure); and accelerating deep learning impact where cloud inference is infeasible.

\paragraph{Limitations, Next Steps, and Code Availability.}
\textbf{Baselines.} Experiment~1 compares SPRKD against response KD (logit matching) and scratch-trained controls on malaria; we do not include feature-based, hint-based, CRD, or relational KD baselines \citep{tung2019similarity, park2019relational}. Broader claims about ``KD methods'' should be read in the context of replication-style distillation unless extended in future work. \textbf{Ablations.} We do not isolate the marginal contributions of ASR initialization, NHE, and PGD perturbations; reviewers may reasonably ask whether NHE alone explains most of the gain---this remains an open empirical question. \textbf{Experiment~2.} TinyImageNet ResNet results are qualitative only (Section~\ref{sec:results}); a full numeric table is deferred. \textbf{Architecture.} TLI with the embedding principle requires matched depth and narrower student width \citep{zhang2021embedding}; this is a real deployment constraint, not merely a notational convenience, and does not automatically extend to arbitrary deep networks (for example, ResNets with residual shortcuts) without careful layer pairing. \textbf{Theory.} A proof of convergence for the combined ASR + NHE + PGD optimizer remains future work. Do findings generalize across non-image tasks and non-SGD optimizers? Can SPRKD be combined with online or hint-based KD? Planned next steps include ablations, additional datasets and transformers, deployment case studies, and comparisons to Born Again Networks \citep{furlanello2018born}, Relational KD \citep{park2019relational}, and Similarity-Preserving KD \citep{tung2019similarity}. Code and reproduction scripts: \url{https://github.com/thetechdude124/SADDLE-POINT-RECRUITMENT-FOR-KNOWLEDGE-DISTILLATION}.

\section*{Acknowledgements}
We thank the SciComm Viewpoint Challenge organizers, the Team Canada ISEF delegation, and reviewers of earlier versions. Experiments used a single Google Colab Tesla GPU and an Intel i3 laptop. This work received 2nd Place in Mathematical and Cybersecurity Research (NSA) and 4th Place in Robotics and Intelligent Machines (Regeneron) at ISEF 2023, and was selected to represent Canada by Youth Science Canada.

\enlargethispage{4\baselineskip}
\begingroup
\footnotesize
\setlength{\bibsep}{0pt plus 0.3ex}
\bibliography{refs}
\endgroup

\end{document}